\documentclass[letterpaper, 10pt, conference]{ieeeconf}
\IEEEoverridecommandlockouts
\usepackage{amsmath,amssymb}
\usepackage{booktabs}
\usepackage{graphicx}
\usepackage[table]{xcolor}
\usepackage[breaklinks=true, colorlinks, bookmarks=false]{hyperref}
\hypersetup{colorlinks=true, linkcolor={red}, citecolor={hanpurple}, urlcolor={magenta}}
\definecolor{rblue}{rgb}{0,0.5,1}
\definecolor{awesome}{rgb}{1.0, 0.13, 0.32}
\definecolor{hollywoodcerise}{rgb}{0.96, 0.0, 0.63}
\definecolor{lasallegreen}{rgb}{0.03, 0.47, 0.19}
\definecolor{hanpurple}{rgb}{0.32, 0.09, 0.98}
\definecolor{green(pigment)}{rgb}{0.0, 0.65, 0.31}
\newcommand{\provia}{\textsc{Provia}}
\title{\LARGE \bf PROVIA: Procedure State Tracking for Online Mistake Detection in Egocentric Videos}
\author{Di Wen$^{1,\dagger}$, Kailun Yang$^{2}$, Jimmy Weissert$^{1}$, Luc Maria Scherrer$^{1}$, Cedric Z\"ollner$^{1}$, Ruiping Liu$^{1}$,\\ Yufan Chen$^{1}$, Jiale Wei$^{1}$, Junwei Zheng$^{1}$, and Kunyu Peng$^{1,*}$
\thanks{$^{1}$The authors are with the Karlsruhe Institute of Technology, Karlsruhe 76131, Germany.}
\thanks{$^{2}$The author is with Hunan University, Changsha 410012, China.}
\thanks{$^{\dagger}$First author (email: {\tt\small di.wen@kit.edu}). $^{*}$Corresponding author (email: {\tt\small kunyu.peng@kit.edu}).}%
}
\begin{document}
\bstctlcite{IEEEexample:BSTcontrol}
\maketitle
\thispagestyle{empty}
\pagestyle{empty}
\begin{abstract}
An assistant watching egocentric video should notice a mistake from past frames alone, before the next step begins, and keep working once the person recovers.  A mistake changes the state of the work, so every later step has to be read against what was done rather than against the plan.  The first-mistake protocol that current online methods report on cuts each recording at its first mistake, so a fixed-time rule that never looks at the video is right on every case.  We evaluate on complete trials, where mistakes and recoveries arise naturally, under a validation false-alarm budget and against controls that use timing alone.  \provia{} keeps two records apart: a factual state, a learned summary of the steps each actor performed, mistakes included, and the accepted progress, an exact posterior over the state of an automaton induced from correct demonstrations by Bayesian state merging and over the execution status of each actor.  Procedure-state transitions occur only in the correct-status branch; the mistake and correction branches retain the source state.  A sequential test turns the per-frame mistake probability into alarms.  With one filter and one optimization rule, \provia{} ranks mistakes best among the evaluated controlled baselines on CaptainCook4D, IndustReal, HoloAssist and IMPACT-ego.  At a validation budget of 0.1 false alarms per minute it recalls .154 against .128 on CaptainCook4D and .034 against .015 on HoloAssist, where it leads at every budget.  The pipeline runs at 58--70 frames per second.  The source code is available at \url{https://github.com/Kratos-Wen/PROVIA}.
\end{abstract}

\section{Introduction}
\label{sec:intro}

A mistake in a procedure has no single form.  What counts as one depends on the domain: a dish is spoiled by the wrong ingredient or the wrong order, an assembly by a part seated the wrong way or a tool used on the wrong fastener.  It depends on scale, since some actions unfold slowly enough to be watched and others are over almost as soon as they begin. And it depends on who is acting, because a procedure carried out by two hands gives each of them a different role, and the progress of the work depends on the actions of both.

Seen from a running egocentric camera, the problem is harder still.  The observer holds a growing prefix.  It does not know where the current step began, cannot look ahead, and cannot select a threshold on the recordings it will be judged on.  Two properties of procedures shape the problem.  Independent steps commute, so what is correct after a prefix is a set of continuations rather than one next step.  And a mistake changes the state of the work, so every later step must be read against what was actually done rather than against the plan.  The person does not stop at the mistake; they may repair it, and they may err again.  An alarm triggers an intervention: a spoken correction, a highlighted part, or a robot that holds back the next part.  Each false alarm interrupts the person, so an assistant runs at a false-alarm budget, and an alarm is most useful before the next step begins.

Evaluation has not kept pace with either difficulty.  The protocol that current online methods report on~\cite{flaborea2024prego,seminara2024dtgl} places the mistake at one designated moment, the first, cuts the recording there, and asks which of the method's own segments was last. A procedure that emits two events at fixed times, without reading a frame, is scored as perfectly correct on both of its benchmarks, and most of the test participants are also seen in training.  We evaluate instead on complete trials as they were performed, where mistakes and the repairs that follow them occur where the person made them, under a stated budget of false alarms, against controls that see only elapsed time, and with every threshold fixed on validation.

\provia{} treats the question as inference about the state of the procedure while the video streams in, and it keeps apart the two things a mistake separates: what the person did and how far the procedure has correctly advanced.  A factual state, updated after every completed step, remembers what each actor was observed to do, mistakes and corrections included.  An exact posterior over the state of a procedure automaton and over the execution status of each actor's current step, correct, mistake, or correction, holds the progress that was accepted.  The automaton is induced from correct demonstrations by Bayesian state merging.  Procedure-state transitions occur only in the correct-status branch, while the mistake and correction branches retain the source state, so the next step is read against what was performed rather than against progress the mistake did not make.  One execution status is carried per actor, so a task described for the performer as a whole and a bimanual task described per hand are read by the same model.  A sequential test turns the per-frame mistake probability into alarms at a stated false-alarm budget.  Our contributions are:
\begin{itemize}
  \item an account of why the protocol current online methods report on admits a perception-free solution with perfect F1, and an evaluation of mistake detection on complete trials, under a stated false-alarm budget and against controls that use no perception;
  \item an online model that keeps the execution as performed apart from the progress it accepts: a factual state of the steps each actor performed, and an exact posterior over procedure state and per-actor execution status on an automaton induced from correct demonstrations by Bayesian state merging;
  \item the highest step-level average precision (AP) and area under the ROC curve (AUROC) among controlled online baselines on four benchmarks that differ in domain, in granularity and in how many hands are described, with one filter and one optimization rule, and ablations showing that removing the record of what was performed at inference costs accuracy on all three single-actor benchmarks.
\end{itemize}

\section{Related Work}
\label{sec:related}

\paragraph{Benchmarks for procedural mistakes.} Assembly101 made multi-step procedures observable at scale and recorded corrections beside mistakes~\cite{sener2022assembly101}, and per-step correct, mistake and correction labels followed~\cite{ding2023every}. Later benchmarks widened the definition: deliberate and natural cooking errors~\cite{peddi2024captaincook4d}, execution errors in industrial-like assembly~\cite{schoonbeek2024industreal}, incorrect actions in real-world tasks~\cite{wang2023holoassist}, typed errors in cooking~\cite{lee2024egoped}, and a six-category anomaly taxonomy for bimanual industrial work in which each hand is annotated separately~\cite{wen2026impact,zhang2026impacthoi}.

\paragraph{Offline and online mistake detection.} Offline methods read a segment whose boundaries are given or a recording that has ended: learned prototypes per step~\cite{lee2024egoped}, action-aware reconstruction~\cite{wang2025amnar}, the visual effect an action should leave~\cite{mur2026aem}, task graphs the step is checked against~\cite{lee2025gtg2vid,lee2026axg}, and zero-shot vision--language models~\cite{seminara2606zeprom}.  Online methods read a growing prefix: PREGO and TI-PREGO segment the stream and compare the recognized step with one a language model anticipates~\cite{flaborea2024prego,flaborea2024tiprego}, differentiable task graphs anticipate with a learned prior~\cite{seminara2024dtgl}, MistSense scores a causal window with hand pose~\cite{patsch2025mistsense}, STORM-PSR recognizes completed steps under occlusion~\cite{schoonbeek2025stormpsr}, and MistExit, given a step, learns when it has seen enough of it~\cite{mistexit2026}.  In PREGO and DTGL the recognized steps, mistakes included, become the history that the next prediction conditions on, and in the prototype, effect and task-graph methods the reference is fixed, so a mistake leaves no trace in it.  This work keeps what was performed apart from what was accepted and scores complete trials at a stated false-alarm rate.

\paragraph{Procedural assistance.} An assistant needs the step in progress, whether it was done correctly, and what may follow~\cite{wang2023holoassist,schoonbeek2024industreal}, resting on recognition that stays reliable outside the training vocabulary~\cite{peng2024openset}.  GuideMe~\cite{liu2026guideme} and Plan-Watch-Recover~\cite{kundu2026pwr} pass a running recording to a language model that decides when to speak, but neither states a rate at which speaking is wrong.  \provia{} keeps the procedure itself as the latent variable, induced from demonstrations by Bayesian state merging~\cite{stolcke1992hmm}, and reports detection at a stated false-alarm rate.

\section{Evaluating Online Mistake Detection}
\label{sec:eval}
\label{sec:eval:validity}
\label{sec:eval:protocol}

\noindent\textbf{The first-mistake protocol.}  PREGO and DTGL evaluate on Assembly101-O and EPIC-Tent-O~\cite{flaborea2024prego,seminara2024dtgl}: each test video is cut at its first mistake, the method segments it, the \emph{last predicted segment} is labeled as the mistake, and the F1 of the correct and mistake classes over predicted segments is averaged.  Four measurements show what it rewards. \emph{(i)} Nothing after the first mistake is tested, although 73\% of the 707 mistakes in the full Assembly101 annotations~\cite{ding2023every} follow an earlier one. \emph{(ii)} The method chooses the unit and the label goes to the last one: two fixed-time events per execution, the second labeled a mistake, give average F1 1.000 on both benchmarks with no perception, where DTGL reports .54 and .47~\cite{seminara2024dtgl}. \emph{(iii)} The label is a position prior: length and step-count priors trained on correct executions only reach average F1 .50 on Assembly101-O and .52 on EPIC-Tent-O, above the .33 and .29 reported for PREGO and at the level of DTGL's .54 and .47~\cite{seminara2024dtgl}, and the normalized position of an event alone separates the terminal event with AUROC .971 on EPIC-Tent-O. \emph{(iv)} The split is not participant-disjoint: 35 of 47 Assembly101-O test participants also appear in training.  The metric reimplementation was checked against the released DTGL evaluator~\cite{seminara2024dtgl}; the evaluation below removes the first three on every benchmark and the fourth where participants are identified.

\noindent\textbf{Alarm-level protocol.}  A recording is scored whole, correct executions included.  A method reads frames as they arrive and may raise an alarm at each of its decisions, here the completions of the shared step segmentation (Sec.~\ref{sec:method:observer}); it is told the task and nothing else about the recording.  Every annotated step owns its completion cell, the times closer to its annotated completion than to any other.  An alarm credits a mistake when it is raised at a decision of the method inside that mistake's cell or at the next decision after the last of them; each mistake is credited once, and one alarm may credit two consecutive mistakes.  An alarm inside the cell of a correct step is false even when it also credits the mistake before it, and a mistake whose cell holds no decision counts against recall.  False alarms are counted per minute of correct operation, the recording time outside mistake steps.  Every method's per-segment score passes through the same sequential rule (Sec.~\ref{sec:method:detection}), with the training-fold mistake prevalence in place of $\gamma_i$ where a method has no prior of its own.  For budgets of 0.1, 0.5 and 1 false alarm per minute, each method, seed and budget receives the threshold of highest validation recall among those whose validation rate meets the budget, at which the test recordings are scored.  We report the recall of all mistakes and the recall of the decoded mistakes that follow another decoded mistake of the same execution.  Three controls pass through the same rule without looking at the video: they score a decision by its index, by the time elapsed, or by how often a training step at that index was a mistake.  A method is credited with detection only where it exceeds every control on the benchmark.

\noindent\textbf{Step-level metrics.}  AP and AUROC are computed over the annotated step segments.  Each segment takes the highest mistake score reached, over the causal prefixes, by any predicted segment whose completion falls in its cell; a segment the model never decoded scores zero.  Segments shorter than 0.5\,s are not scored.

\section{Method}
\label{sec:method}

\begin{figure*}[t]
  \centering
  \includegraphics[width=0.90\textwidth]{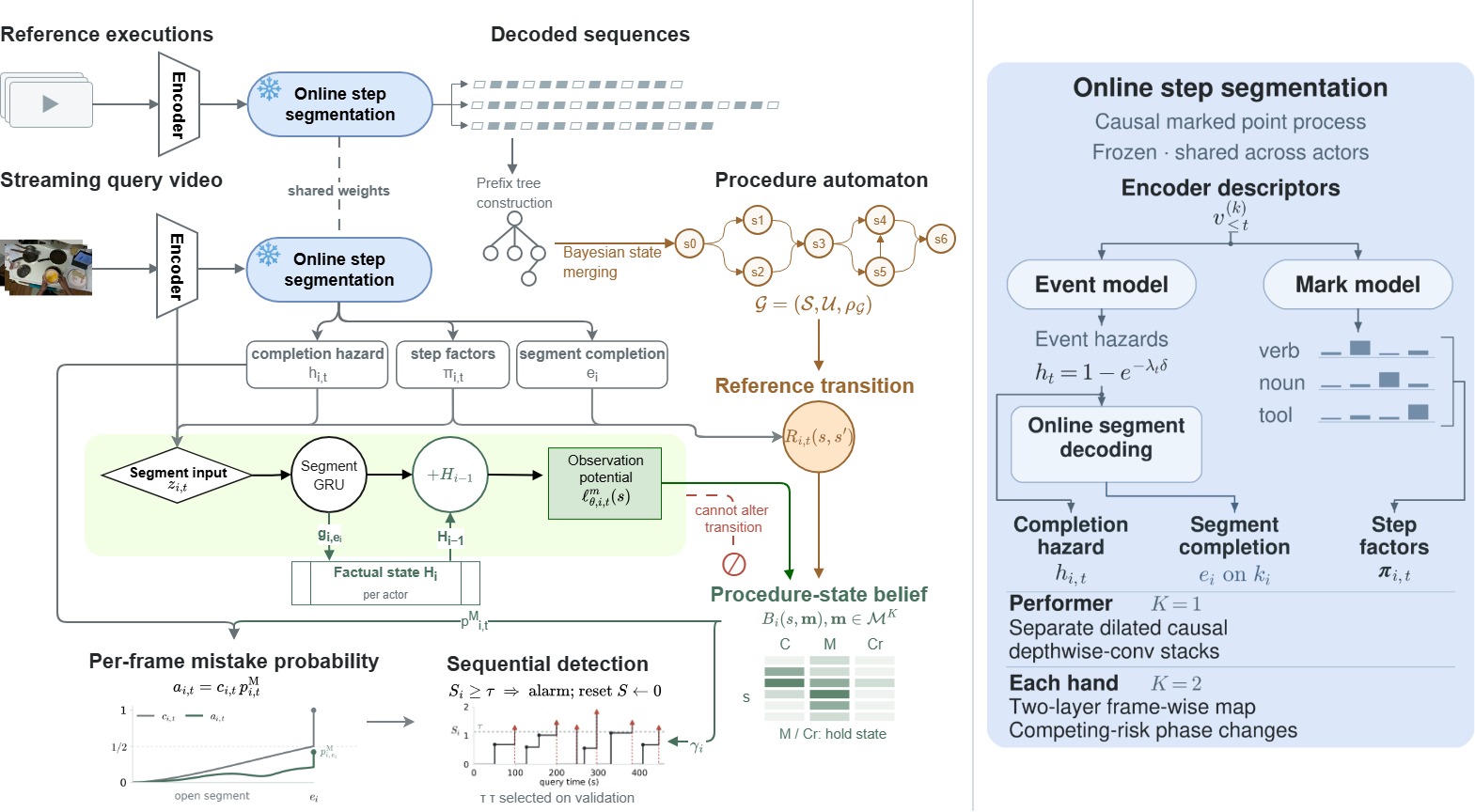}
\caption{\provia{} tracks a procedure carried out by $K$ actors.  Left: correct executions are decoded into actor-tagged step sequences and merged into one automaton by Bayesian state merging; on the query stream, the factual state records the steps each actor performed and an exact filter over procedure state and per-actor execution status, $B_i(s,\mathbf m)$, records the progress accepted.  Right: the segmentation module, shared across actors.}
  \label{fig:overview}
\end{figure*}

One filtering formulation handles one or more actors (Fig.~\ref{fig:overview}).

\subsection{Problem Formulation}
\label{sec:method:problem}

Let $x_{1:T}$ be a query video and let $\mathcal D^{\mathrm{ref}}$ be the set of correct executions of the same task in the training partition; a task without a complete correct recording falls back on the deterministic medoid of its decoded step sequences, without labels.  At query time the model receives only $x_{\leq t}$ and the task of the recording, which selects $\mathcal D^{\mathrm{ref}}$: no query step boundary, step label, correctness label, or future frame.

A procedure is carried out by $K$ actors: $K=1$ when its steps describe the person as a whole, $K=2$ when each hand carries a distinct part of the work.  Each actor $k$ carries a descriptor $v^{(k)}_t$ per frame.  From $v^{(k)}_{\leq t}$ the segmentation model produces, at every frame, soft step factors $\boldsymbol\pi_t=\{\boldsymbol\pi_t^{\mathrm{verb}}, \boldsymbol\pi_t^{\mathrm{noun}},\boldsymbol\pi_t^{\mathrm{tool}}\}$ and a completion hazard $h_t$, and it partitions the actor's observations exhaustively into predicted step segments: a segment closes at the model's decision and the next opens at once, so every actor always carries an open step.  Segments are indexed $i=1,\ldots,I$ in the order of their completions across all actors, an open segment receiving its index at completion; $k_i$ denotes the actor of segment $i$, and $b_i$ and $e_i$ its predicted start and completion.  For a segment observed up to time $t$, the filter returns $p_{i,t}^{\mathrm M}=p(m_i=\mathrm M\mid x_{\leq t},\mathcal D^{\mathrm{ref}})$, where $m_i\in\mathcal M=\{\mathrm C,\mathrm M,\mathrm{Cr}\}$ is the execution status of the segment: correct, mistake, or correction.  Where the segmentation model can propose a segment that belongs to no step of the procedure, $\mathcal M$ carries a fourth status, insertion (Ins).  The reported output is $a_{i,t}=c_{i,t}\,p_{i,t}^{\mathrm M}$, where the completion gate $c_{i,t}\in[0,1]$ is the mass of the segment's status that has been decided by frame $t$ (Sec.~\ref{sec:method:detection}).

\subsection{Online Step Segmentation}
\label{sec:method:observer}

For every actor the segmentation model is the same causal marked point process with shared parameters: its event types follow the annotation and its marks are the step labels.  A causal network reads the descriptor stream and outputs, at every frame, an intensity $\lambda_t$ per event type (softplus, offset by the fold base rate) and the step factors $\boldsymbol\pi_t$ through one head per factor, giving the hazard $h_t=1-e^{-\lambda_t\delta}$ for the frame interval $\delta$.  Training minimizes the point-process negative log-likelihood of the annotated event times, piecewise constant over frames, plus the step-label negative log-likelihood corrected by the Jeffreys class prior of the fold; mistake labels are not used, and the event and mark branches are selected separately by their validation likelihood.  Decoding is the causal posterior median of the next event time: an event is decoded at the first frame at which the survival $\prod_{\nu}(1-h_\nu)$ accumulated since the preceding decoded event falls to one half, with no score, duration, gap, or suppression threshold, and the same rule forms segments from events under either annotation.  When steps are annotated as intervals of the performer there is one event type, step completion, and every decoded event closes the open segment and opens the next.  When they are annotated per hand, the events are the phase changes of that hand among idle, approach and interaction, with one intensity per change as competing risks; a hand's segment opens when it leaves idle or renews directly out of an interaction, its mark is fixed at the interaction onset, and it closes when the hand leaves the interaction.

\subsection{Procedure Automaton Induction}
\label{sec:method:language}

Because correct executions differ in local order and one path per execution memorizes the training set, we induce a probabilistic finite automaton from the decoded demonstrations $\mathcal D^{\mathrm{ref}}$.  Each demonstration is decoded on all $K$ of its actors, and the decoded steps are interleaved by completion time into one sequence of actor-tagged steps $(k,u)$; the automaton is the prefix tree of these sequences, keeping every observed branch and revisit.  With $K=2$ the order of the two hands' steps is part of the language.

The tree is then reduced by Bayesian state merging~\cite{stolcke1992hmm}: two histories share a state when they predict the same continuation law. For a state $s$, let $\mathbf n_s$ count termination and the next step label, with $N_s=\sum_u n_{s,u}$.  Its integrated evidence $\mathcal E(\mathbf n_s)$ is the product of a Jeffreys Beta--Bernoulli evidence for termination and a Jeffreys--Dirichlet evidence for the next step label.  Two incomparable states $s_1$ and $s_2$ are pooled iff
\begin{equation}
    \Delta_{s_1,s_2}=\log\mathcal E(\mathbf n_{s_1}+\mathbf n_{s_2})
    -\log\mathcal E(\mathbf n_{s_1})-\log\mathcal E(\mathbf n_{s_2})>0.
    \label{eq:quotient-bayes-factor}
\end{equation}
Equal prior odds decide whether one shared next-step law is better supported than two, without a similarity threshold, state count, or edit cost.  Merging is greedy over pairs that share an observed next step, largest $\Delta$ first with rescoring, until no pair has $\Delta>0$; ties break by state index, and ancestor and descendant states are never merged, so repeats and order-flexible behavior stay representable.

The resulting automaton $\mathcal G=(\mathcal S,\mathcal U,\rho_{\mathcal G})$ consists of procedure states $s\in\mathcal S$, admissible next steps $u\in\mathcal U(s)$, and a posterior-predictive joint law $\rho_{\mathcal G}(u,s'\mid s)$.  A continuation absent from the $n^{\mathrm{ref}}$ correct executions keeps the Jeffreys posterior-predictive probability $\varepsilon=\tfrac12/(n^{\mathrm{ref}}+1)$.  The demonstrations also say what each step changes.  Let $\mathbf o_i=v^{(k_i)}_{e_i}-v^{(k_i)}_{b_i}$ be the change of the actor's descriptor over segment $i$, standardized on the demonstrations.  A diagonal Gaussian random-effects model, with within- and between-step variances estimated by moments, gives step $u$ a posterior mean $\boldsymbol\mu_u$ shrunk toward the task mean and a predictive variance $\boldsymbol\Sigma_u$, and the step effect of segment $i$ is
\begin{equation}
    \beta_i(u)=\log\mathcal N(\mathbf o_i;\boldsymbol\mu_u,\boldsymbol\Sigma_u)-\log\sum_{u'}\omega_{u'}\,\mathcal N(\mathbf o_i;\boldsymbol\mu_{u'},\boldsymbol\Sigma_{u'}),
    \label{eq:step-effect}
\end{equation}
with log densities averaged over coordinates, $\omega_u$ the Jeffreys frequency of $u$ among the demonstrated steps, and $\beta_i(u)=0$ before completion.  The automaton and the step effects use no annotated boundary or label and no validation or test video.

\subsection{Bayesian Filtering over Procedure State}
\label{sec:method:observation}
\label{sec:method:transition}

Segment $i$ owns a bounded recurrent state, and the factual state is updated when the segment completes:
\begin{equation}
\begin{aligned}
 z_{i,t}&=\operatorname{LN}\!\left(\phi_{\mathrm c}(v_{\leq t})+
 \textstyle\sum_f E_f^{\top}\boldsymbol\pi_{i,t}^{f}+\psi(h_{i,t})\right),\\
 g_{i,t}&=\operatorname{GRU}_{\theta}(g_{i,t-1},z_{i,t}),\qquad
 q_{i,t}=\operatorname{LN}(g_{i,t}+H_{i-1}),\\
 H_i&=\operatorname{GRU}'_{\theta}\bigl(H_{i-1},\,W(g_{i,e_i}+\textstyle\sum_f E_f^{\top}\boldsymbol\pi_{i,e_i}^{f})\bigr),
\end{aligned}
\label{eq:candidate-state}
\end{equation}
where LN is layer normalization, GRU a gated recurrent unit, $\phi_{\mathrm c}$ a causal window encoder trained with the classifier, $\psi$ a map of the hazard, $E_f$ the embedding of step factor $f$, $W$ a linear map, and $g_{i,t}$ runs over the frames of segment $i$ only.  $H_i$ is updated after every segment, whatever its execution status, so it records what was observed.  Each actor owns a factual state; the query of a segment on actor $k$ adds the factual state of $k$, a linear map of those of the other actors, and an actor embedding, which with $K=1$ reduces to $H_{i-1}$.

Given a procedure state $s$, the classifier produces a status energy for each $m\in\mathcal M$ from a key $\mathbf k_s$ that embeds three automaton statistics of $s$ (successor support, termination probability, continuation entropy), and the observation potential of the segment under state $s$ and status $m$ is the log-softmax
\begin{equation}
    \ell^{m}_{\theta,i,t}(s)=\log\operatorname{softmax}_m\Bigl[b_m(q_{i,t})+\tfrac{1}{\sqrt d}\,\mathbf k_s^{\top}W_m q_{i,t}\Bigr],
    \label{eq:observation-likelihood}
\end{equation}
with a bias $b_m$ and a projection $W_m$ per status and $d$ the state dimension.

Let $\mathbf m=(m^{(1)},\ldots,m^{(K)})\in\mathcal M^K$ collect the execution status of the current segment of every actor, and let $B_{i-1}(s,\mathbf m)$ be the posterior over procedure state and status vector at the boundary of segment $i$.  $H_i$ records what was observed; $B_i$ records which progress was accepted.  Under the mistake and correction branches $B$ keeps its state while $H$ carries the segment, so the next segment, and a correction in particular, is read against what was performed and not against a state the mistake would have created.  The automaton defines the reference transition
\begin{equation}
    R_{i,t}(s,s')=\sum_{u\in\mathcal U(s)}
    \rho_{\mathcal G}(u,s'\mid s)\,
    p_{\mathcal G}(\boldsymbol\pi_{i,t}\mid u)\,
    \exp\beta_i(u),
    \label{eq:reference-transition}
\end{equation}
where $p_{\mathcal G}(\boldsymbol\pi_{i,t}\mid u)=\prod_f|\mathcal Y_f|\,\langle\boldsymbol\pi^{f}_{i,t},\boldsymbol\alpha^{f}_{u}\rangle$ is the likelihood ratio of the observed step factors under the label law $\boldsymbol\alpha^f_u$ of step $u$ against a uniform law over the labels $\mathcal Y_f$; it equals one when the segmentation model is uninformative.  A step the segmentation model misses, independently with the training-fold miss rate $\kappa<1$, is absorbed by the closure $\Gamma=(I-\kappa\bar\rho_{\mathcal G})^{-1}\operatorname{diag}(\mathbf 1-\kappa\bar\rho_{\mathcal G}\mathbf 1)$, whose rows sum to one, where $\bar\rho_{\mathcal G}(s'\mid s)=\sum_u\rho_{\mathcal G}(u,s'\mid s)$; a continuation absent from the demonstrations enters with probability $\varepsilon$ and leaves the state unchanged.  The correct-status operator is
\begin{equation}
    \Lambda^{\mathrm C}_{i,t}(s,s')=
    e^{\ell_{\theta,i,t}^{\mathrm C}(s)}\bigl[\varepsilon\,\mathbf 1[s'=s]+(1-\varepsilon)(\Gamma R_{i,t})(s,s')\bigr].
    \label{eq:protected-transition}
\end{equation}
The classifier does not enter the bracket.  Its factor $e^{\ell^{\mathrm C}_{\theta,i,t}(s)}$ is constant across the destinations of a source state and cancels when a row is normalized, so the classifier reweights which source states and statuses are believed but cannot choose where accepted progress moves.

Mistake, correction and insertion explain the observation without advancing the procedure state: $\Lambda^{m}_{i,t}(s,s')=\mathbf 1[s'=s]\exp \ell_{\theta,i,t}^{m}(s)$ for $m\neq\mathrm C$. At the predicted completion of segment $i$ on actor $k=k_i$, the exact filtering update acts on the procedure state and on the status coordinate of that actor alone,
\begin{equation}
\begin{aligned}
    \widetilde B_i(s',\mathbf m')
    &=\sum_{s,\,m^{(k)}}B_{i-1}\bigl(s,\mathbf m'_{[k\leftarrow m^{(k)}]}\bigr)
      P_{\mathrm{tr}}\bigl(m'^{(k)}\mid m^{(k)}\bigr)\\[-1mm]
    &\hspace{17mm}\times\Lambda^{m'^{(k)}}_{i,e_i}(s,s'),
\end{aligned}
    \label{eq:state-update}
\end{equation}
and $B_i\propto\widetilde B_i$, where $\mathbf m'_{[k\leftarrow m]}$ replaces the $k$-th coordinate of $\mathbf m'$ by $m$, the admissible steps in $\Lambda$ are those tagged with actor $k$, and $P_{\mathrm{tr}}$ and the initial status law are the Jeffreys posterior-predictive estimates (counts plus one half) from the status sequences of the training fold.  With $K=2$ the recursion runs for whichever hand completes; when both complete on the same frame the two orders of application are averaged.  With $K=1$, Eq.~\eqref{eq:state-update} is applied once, at $e_i$; before $e_i$ the segment likelihood changes the reported posterior but not the procedure state.  With $K=2$ the belief also carries, for each open segment, the status mass that the clock of Sec.~\ref{sec:method:detection} has not yet decided: the fraction decided at a frame passes through Eq.~\eqref{eq:state-update} with the evidence of that frame, so a fraction decided as correct advances the procedure state at its decision frame, every decided branch keeps receiving the later evidence of the segment, and at $e_i$ the remaining mass is decided and the branches are merged into $B_i$.  In both cases each unit of status mass passes through the update once, and the procedure state advances only along the correct-status branch.

\subsection{Sequential Detection Rule}
\label{sec:method:detection}

The status of an open segment starts unresolved, and a clock moves its mass to decided frame by frame; the completion gate $c_{i,t}$ is the decided mass, and at the predicted completion the remaining unresolved mass is decided, so $c_{i,e_i}=1$ under either annotation.  When steps describe the performer, the clock is the completion posterior induced by the hazard, $c_{i,t}=1-\prod_{\nu=b_i}^{t}(1-h_{i,\nu})$.  When they describe a hand, the segmentation model supplies the causal posterior that the hand's interaction has begun, which fixes the segment's identity, and the clock is a two-state head on $q_{i,t}$, learned with the observation model, whose input includes that posterior.  The mistake probability reported at frame $t$ is $a_{i,t}=c_{i,t}\,p(m_i=\mathrm M\mid x_{\leq t},B_{i-1})$: it rises with the evidence that the step is complete and with the evidence that it was wrong.  No latency loss or time threshold enters it.

Let $\gamma_i=\sum_{s,\mathbf m}B_{i-1}(s,\mathbf m)\,P_{\mathrm{tr}}(\mathrm M\mid m^{(k_i)})$ be the probability that the filter assigns to a mistake in segment $i$ before the segment is observed, and $a_i=p(m_i=\mathrm M\mid x_{\leq e_i},B_{i-1})$ the posterior at its completion.  The ratio of the posterior odds to the prior odds
\begin{equation}
    \xi_i=\frac{a_i/(1-a_i)}{\gamma_i/(1-\gamma_i)}
    \label{eq:bayes-factor}
\end{equation}
is the evidence score of segment $i$; the division normalizes the score by the status prior that $P_{\mathrm{tr}}$ has already propagated.  The Shiryaev--Roberts statistic~\cite{shiryaev1963optimum,roberts1966comparison} $S_i=(1+S_{i-1})\,\xi_i$, $S_0=0$, accumulates these scores and equals $\sum_{r\leq i}\prod_{j=r}^{i}\xi_j$, the sum over onsets $r$ of the products of the scores from $r$ to $i$.  An alarm is raised when $S_i\geq\tau$ and resets $S$ only; the filter and the factual state are unchanged.  The threshold $\tau$ is set for a stated false-alarm budget (Sec.~\ref{sec:eval:protocol}).

\subsection{Learning Objective}
\label{sec:method:learning}

The segmentation model is trained first and frozen, the automaton and the step effects are extracted from its decodes by Eqs.~\eqref{eq:quotient-bayes-factor}--\eqref{eq:step-effect}, and the observation model is optimized on the frozen segments, so mistake supervision can neither move segment boundaries nor rewrite the automaton.  Each predicted segment inherits the status of the annotated step it is matched to.  With $K=1$, the predicted segment nearest to an annotated completion inside its completion cell inherits that step's status, one-to-one, and the other segments are correct context.  With $K=2$, the segments of each hand are matched to that hand's annotated steps by a maximum-cardinality monotone matching of overlapping intervals, and an unmatched segment is an insertion.

Let $p_{i,m}$ be the end-of-segment posterior for execution status $m$, $\widehat\omega_m=(n_m+\tfrac12)/(\sum_{m'}n_{m'}+\tfrac{|\mathcal M|}{2})$ the fold-training Jeffreys status law, and $w_m\propto\widehat\omega_m^{-1}$ its inverse-prior weight.  We use the prior-preserving case-control log score
\begin{equation}
    \bar p_{i,m}=\frac{w_mp_{i,m}}{\sum_{m'}w_{m'}p_{i,m'}},
    \qquad
    \mathcal L=-\frac{\sum_iw_{y_i}\log\bar p_{i,y_i}}
                         {\sum_iw_{y_i}}.
    \label{eq:training-objective}
\end{equation}
With $K=2$ the same score trains the clock: the end-of-segment posterior is a mixture over the frames at which the clock decided the status mass (Sec.~\ref{sec:method:observation}), and $\mathcal L$ reaches the clock through this mixture; no separate loss or latency term is used.  When a benchmark does not annotate corrections, the likelihood marginalizes $\mathrm C$ and $\mathrm{Cr}$, so a correction is never relabeled as a mistake.

\section{Experiments}
\label{sec:experiments}

\subsection{Setup}
\label{sec:experiments:protocol}

\paragraph{Datasets.} We use four public egocentric benchmarks with complementary kinds of mistakes: CaptainCook4D~\cite{peddi2024captaincook4d}, IndustReal~\cite{schoonbeek2024industreal}, HoloAssist~\cite{wang2023holoassist}, and the egocentric view of IMPACT~v1.1~\cite{wen2026impact}.  CaptainCook4D and IndustReal use their official participant-disjoint splits; for HoloAssist, whose test labels are not released, a video-group-disjoint validation subset is carved from the official training partition and the official validation partition is the outer test (1,204/262/207 recordings); IMPACT's official splits are not participant-disjoint, so we use a deterministic participant-disjoint fold (57/15/21 executions from 8/2/3 participants). The outer test sets contain 442/1,391, 17/194, 931/17,705, and 177/1,760 mistakes/step segments (prevalence .318, .088, .053, .101).  Every method is trained with three seeds, and results are means over them.

\begin{table}[t]
\centering
\caption{Classes are counted over the whole corpus, Granularity is the median interval between the step completions of one actor, Steps/exec is the median number of steps per execution, and Demos/task is the mean number of error-free training executions per task.}
\label{tab:benchmarks}
\footnotesize
\resizebox{\linewidth}{!}{\begin{tabular}{llrrrrrrc}
\toprule
Benchmark & Domain & h & Steps & Classes & Steps/exec & Demos/task & Granularity & $K$ \\
\midrule
CaptainCook4D~\cite{peddi2024captaincook4d} & cooking & 94 & 5,394 & 360 & 14 & 3.9 & 43.3\,s & 1 \\
IndustReal~\cite{schoonbeek2024industreal} & assembly & 6 & 501 & 39 & 6 & 12.0 & 38.3\,s & 1 \\
HoloAssist~\cite{wang2023holoassist} & mixed tasks & 122 & 141,212 & 2,197 & 63 & 10.8 & 1.9\,s & 1 \\
IMPACT-ego~\cite{wen2026impact} & disassembly & 6 & 8,015 & 134 & 79 & 15.0 & 2.7\,s & 2 \\
\bottomrule
\end{tabular}}
\end{table}

\paragraph{Shared feature streams.} Each benchmark has one feature setting shared by every method.  On the three single-actor benchmarks it is the frozen DINOv2 ViT-S/14 encoder~\cite{oquab2023dinov2} on $224\!\times\!224$ RGB frames, sampled at 8~Hz.  On IMPACT-ego it is an in-domain hand-object stream per hand, since frozen features leave every controlled system at chance there (AUROC .45--.49): a YOLOv9-c detector (640\,px, confidence .25) finds each hand at 15~Hz, a crop enlarged 1.6 times around it is encoded by a DINOv2-S fine-tuned on training-fold step labels only, and the stream is that descriptor and its rate of change.  The segmentation model of Sec.~\ref{sec:method:observer} runs on these streams, so every method receives the same segments, hazard, and step factors.

\subsection{Implementation Details}
\label{sec:experiments:implementation}

\paragraph{Segmentation model.} On the three single-actor benchmarks ($K=1$) the event is step completion; frame $t$ enters at 8~Hz as $[v_t,\,v_{t-1},\,\dot v_t]$, and each branch is a stack of dilated causal depthwise convolutions (width 192, kernel 3, dilations $1,\ldots,2^{L-1}$), with $L$ the smallest value whose causal support covers the longest gap between consecutive training completions (4.3--17~min); 9.4--11.2M parameters.  On IMPACT-ego ($K=2$) the phase changes are fitted with an exact interval-censored likelihood, and the mark is read from the start of the approach to the onset of the interaction.  IMPACT annotates no contact onset, so the interaction onset of a training step is taken at the midpoint of its annotated interval; the network is a frame-wise two-layer map (1.5M parameters), unmarked intervals become segments with the prior step distribution, and the insertion status of Sec.~\ref{sec:method:problem} is used. The segmentation model is trained with AdamW (weight decay $10^{-2}$, gradient clipping 1.0) and early stopping on the validation likelihood, with the learning rate of each branch selected from $\{10^{-4},3\!\times\!10^{-4},10^{-3}\}$ by the same criterion.

\paragraph{Filter.} The causal window encoder $\phi_{\mathrm c}$ of Eq.~\eqref{eq:candidate-state} is a linear map followed by six dilated causal convolutions with a 16\,s support, and $\psi$ is a single linear layer. \provia{} uses a 192-dimensional segment and factual state and one optimization rule on all datasets (AdamW, $3\!\times\!10^{-4}$, weight decay $10^{-2}$, at most 80 epochs, early stopping on the validation log score); AP, AUROC, thresholds, and latency never enter checkpoint selection.  It has 0.71--0.78M trainable parameters and no per-dataset head, task embedding, or language model. One NVIDIA A100 at batch one takes 13.4--16.1~ms per RGB frame for encoder and segmentation model, and the exact update adds at most 0.9~ms, so the pipeline runs at 58--70 frames per second.  On the same stream the scoring heads of the baselines below add 1.0~ms (Causal-TCN), 1.1--1.3~ms (MistSense-RGB-style) and 0.6--12.5~ms (Controlled PREGO, whose window is re-read at every frame) per frame.

\paragraph{Baselines.}
\label{sec:experiments:baselines}
As released, PREGO queries a language model at every step and MistSense reads hand pose, both were scored on recordings cut at the first mistake, and neither returns a score with a stated false-alarm rate.  We therefore compare with three online alternatives that receive the same stream and differ in how they carry the procedure: \emph{Causal-TCN}, a supervised visual control of six left-padded temporal blocks; \emph{Controlled PREGO}, the official MiniROAD~\cite{an2023miniroad} recognizer of PREGO~\cite{flaborea2024prego} with its language-model call replaced by a deterministic anticipator learned from correct training prefixes; and \emph{MistSense-RGB-style}, the RGB Video-Q-Former path of MistSense~\cite{patsch2025mistsense} without pose or its explanation-only language model.  All use the seeds of \provia{}, are trained with their published objective and selected on validation loss, derive temporal windows once from correct training durations, and read the same predicted segments and past frames only.  Controlled PREGO is scored by one minus the posterior mass of the anticipated successor, a continuous score in place of its binary mismatch decision.

\subsection{Comparison with Controlled Baselines}
\label{sec:experiments:main}

\begin{table*}[t]
\centering
\caption{Step-level results of all methods, as mean $\pm$ SD over three training seeds; bold marks the best value. On IMPACT-ego ($^\dagger$) all methods share an in-domain hand-object stream.}
\label{tab:main-results}
\footnotesize
\resizebox{\textwidth}{!}{\begin{tabular}{lcccccccc}
\toprule
& \multicolumn{2}{c}{CaptainCook4D} & \multicolumn{2}{c}{IndustReal} & \multicolumn{2}{c}{HoloAssist} & \multicolumn{2}{c}{IMPACT-ego$^\dagger$} \\
\cmidrule(lr){2-3}\cmidrule(lr){4-5}\cmidrule(lr){6-7}\cmidrule(lr){8-9}
Method & AP $\uparrow$ & AUROC $\uparrow$ & AP $\uparrow$ & AUROC $\uparrow$ & AP $\uparrow$ & AUROC $\uparrow$ & AP $\uparrow$ & AUROC $\uparrow$ \\
\midrule
Causal-TCN & .336$\pm$.007 & .527$\pm$.014 & .148$\pm$.040 & .616$\pm$.055 & .111$\pm$.012 & .633$\pm$.002 & .127$\pm$.009 & .546$\pm$.008 \\
Controlled PREGO~\cite{flaborea2024prego} & .330$\pm$.003 & .496$\pm$.007 & .133$\pm$.010 & .612$\pm$.008 & .057$\pm$.002 & .540$\pm$.007 & .086$\pm$.002 & .442$\pm$.006 \\
MistSense-RGB-style~\cite{patsch2025mistsense} & .346$\pm$.006 & .522$\pm$.018 & .101$\pm$.007 & .567$\pm$.028 & .105$\pm$.009 & .625$\pm$.008 & .134$\pm$.009 & .548$\pm$.010 \\
\provia{} & \textbf{.364$\pm$.024} & \textbf{.554$\pm$.018} & \textbf{.153$\pm$.029} & \textbf{.625$\pm$.022} & \textbf{.131$\pm$.011} & \textbf{.670$\pm$.009} & \textbf{.147$\pm$.008} & \textbf{.567$\pm$.007} \\
\bottomrule
\end{tabular}}
\end{table*}

\noindent\textbf{Step-level results.}  \provia{} has the highest AP and AUROC among the evaluated controlled baselines on every benchmark (Table~\ref{tab:main-results}).  The margin over the best baseline is largest on HoloAssist, the single-actor benchmark with the most steps per execution, where resetting the record of what was performed costs the largest share of the AP, a seventh (Sec.~\ref{sec:experiments:ablation}).

\noindent\textbf{IMPACT-ego.}  Step identity limits every method there: its steps last one to two seconds and the two hands interleave.  With segments only where the model decodes an event, 57\% of the annotated steps hold a decision; keeping an open step on every hand raises this to 86\% and lifts \provia{} from .129 to .147 AP.

\subsection{Ablation Studies}
\label{sec:experiments:ablation}

Table~\ref{tab:ablation-main} removes one component at a time from the frozen model at inference.  \emph{w/o factual state} resets $H_i$ to its initial value before every segment, \emph{w/o status transition} replaces $P_{\mathrm{tr}}$ by the training-fold initial law, \emph{w/o procedure-state belief} replaces the posterior by a uniform law over the states with an admissible successor, and \emph{w/o completion gate} reports the marginal prefix mistake probability.

\begin{table}[t]
\centering
\caption{Ablation of the frozen model at inference: test AP as means over three training seeds. The retrained variant is a matched architecture trained from scratch without the factual state.}
\label{tab:ablation-main}
\footnotesize
\resizebox{\linewidth}{!}{\begin{tabular}{lccc}
\toprule
& \multicolumn{3}{c}{AP $\uparrow$} \\
\cmidrule(lr){2-4}
Variant & HoloAssist & IndustReal & CaptainCook4D \\
\midrule
\provia{} (full) & .131 & .153 & .364 \\
\midrule
w/o factual state $H_i$ & .113 & .138 & .328 \\
w/o factual state $H_i$ (retrained) & .121 & .159 & .333 \\
w/o status transition $P_{\mathrm{tr}}$ & .127 & .135 & .337 \\
w/o procedure-state belief & .127 & .147 & .371 \\
w/o completion gate & .123 & .119 & .370 \\
\bottomrule
\end{tabular}}
\end{table}

\noindent\textbf{Factual state and status transition.}  Resetting the factual state before every segment costs AP on all three benchmarks, a seventh of it on HoloAssist and a tenth on IndustReal and CaptainCook4D, and AUROC moves with it (.670 to .653 on HoloAssist, .554 to .516 on CaptainCook4D).  Retrained from scratch without $H_i$ at the same parameter count, the model stays below the full one on HoloAssist and CaptainCook4D and above it on IndustReal, so on the two larger benchmarks the parameter count does not account for the gain.  The status transition is the second dependence on the past, a prior on how the next execution status follows the last; removing it costs AP everywhere and most on IndustReal and CaptainCook4D.

\noindent\textbf{Belief over procedure states.}  The belief contributes little at the step level.  The induced automata are nearly linear: after merging, a state admits 1.01 to 1.18 next steps on average and the merge compresses the prefix tree of the demonstrations by 1\% to 14\%, and a belief over a nearly linear automaton carries little beyond position.

\noindent\textbf{Completion gate and CaptainCook4D.}  The gate ties the reported probability to the evidence that the step has ended and is worth most on IndustReal.  On CaptainCook4D the gate and the belief do not help and the status transition carries the benchmark.  With .318 of its steps mistaken, the highest prevalence of the four, only 3.9 executions of a recipe are error-free from end to end, so the automaton is a prefix tree of about four traces over a 27.6-step sequence, whereas the status transition is estimated from every training execution.

\begin{table*}[t]
\centering
\caption{Alarm-level results with thresholds frozen on validation, as means over three training seeds. Columns 0.1, 0.5 and 1.0 give the recall of all mistakes at validation budgets of that many false alarms per minute of correct operation, FA the false-alarm rate per minute reached at the 0.1 threshold, and Later the recall at 0.5 of the decoded mistakes that follow another decoded mistake.}
\label{tab:alarm_test}
\footnotesize
\resizebox{\textwidth}{!}{\begin{tabular}{lccccc|ccccc|ccccc}
\toprule
& \multicolumn{5}{c|}{CaptainCook4D} & \multicolumn{5}{c|}{IndustReal} & \multicolumn{5}{c}{HoloAssist} \\
Method & 0.1$\uparrow$ & FA$\downarrow$ & 0.5$\uparrow$ & 1.0$\uparrow$ & Later$\uparrow$ & 0.1$\uparrow$ & FA$\downarrow$ & 0.5$\uparrow$ & 1.0$\uparrow$ & Later$\uparrow$ & 0.1$\uparrow$ & FA$\downarrow$ & 0.5$\uparrow$ & 1.0$\uparrow$ & Later$\uparrow$ \\
\midrule
Causal-TCN & .100 & .094 & .600 & \textbf{.845} & \textbf{.709} & \textbf{.255} & .126 & \textbf{.529} & \textbf{.882} & \textbf{1.00} & .015 & .094 & .044 & .099 & .058 \\
Controlled PREGO~\cite{flaborea2024prego} & .101 & .105 & \textbf{.605} & .825 & .694 & .137 & .103 & .392 & .863 & .667 & .009 & .095 & .044 & .087 & .058 \\
MistSense-RGB-style~\cite{patsch2025mistsense} & .128 & .103 & .604 & .805 & .698 & .118 & .112 & \textbf{.529} & .745 & \textbf{1.00} & .008 & .102 & .048 & .095 & .063 \\
\provia{} & \textbf{.154} & .075 & .600 & .839 & .701 & .098 & .126 & .392 & .667 & .444 & \textbf{.034} & .093 & \textbf{.124} & \textbf{.210} & \textbf{.126} \\
\midrule
Position control (index) & .090 & .099 & .473 & .826 & .599 & .176 & .129 & .529 & .882 & 1.00 & .000 & .000 & .041 & .078 & .054 \\
Position control (time) & .102 & .117 & .532 & .765 & .676 & .294 & .129 & .588 & .765 & 1.00 & .014 & .061 & .055 & .071 & .071 \\
Position control (training) & .100 & .108 & .430 & .656 & .507 & .176 & .138 & .529 & .765 & 1.00 & .018 & .103 & .058 & .139 & .068 \\
\bottomrule
\end{tabular}}
\end{table*}

\subsection{Qualitative Results}
\label{sec:experiments:qualitative}

\begin{figure}[t]
\centering
\includegraphics[width=\linewidth]{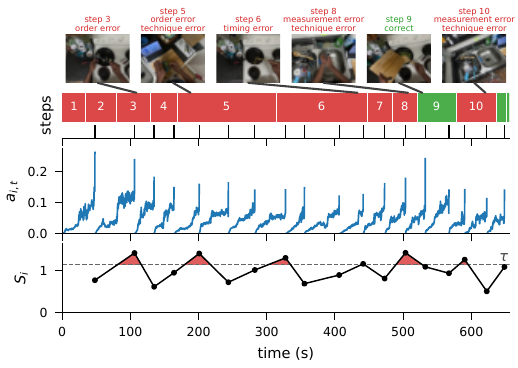}
\caption{One CaptainCook4D test execution (Zoodles, 656\,s) with the seed-0 model. Top: frames from six of the steps, three seconds before their completion, with the step index and the annotated error types above each frame and a line to the frame time. Second row: the twelve annotated steps as their completion cells, red for mistakes and green for correct steps, with the eighteen predicted completions as ticks. Third row: the reported mistake probability $a_{i,t}$ at every frame. Bottom: the Shiryaev--Roberts statistic $S_i$ at each completion against the threshold $\tau$ frozen on validation at 0.5 false alarms per minute; the shaded regions above $\tau$ are the alarms.}
\label{fig:qualitative}
\end{figure}

Fig.~\ref{fig:qualitative} follows one execution through the detector.  The six alarms all fall inside mistake cells, two in the same cell; five credit the mistake of their own cell and three of these also the preceding mistake, whose cell held decisions but no alarm, so eight of the nine mistakes are credited; the first, at 23\,s, has no decision in its cell, since the first segment completes at 48\,s.

\subsection{Detection under a False-Alarm Budget}
\label{sec:experiments:budget}

An assistant is operated at a false-alarm budget, and Table~\ref{tab:alarm_test} scores every method's alarms on the three single-actor benchmarks under the protocol of Sec.~\ref{sec:eval:protocol}.

\noindent\textbf{Alarm-level results across benchmarks.}  At 1 alarm per minute the position controls already recover most of what any method recovers on CaptainCook4D; the methods separate at the strict budgets.  On HoloAssist \provia{} leads at every budget and does so at the lowest measured false-alarm rate of the four methods; on CaptainCook4D it leads at 0.1 per minute at the lowest measured rate of any row, and at 0.5 per minute it reaches the recall of Causal-TCN with fewer false alarms.  The lead at 0.1 rests on the factual state: with $H_i$ reset before every segment, the CaptainCook4D recall falls from .154 to .112.  The order does not depend on the grace of one decision: with one-to-one crediting, where an alarm credits only the mistake of its own cell, every recall falls, and \provia{} still leads on HoloAssist at every budget (.023, .090 and .151 against at most .010, .040 and .087 for any other row) and on CaptainCook4D at 0.1 per minute (.094 against .081).  On IndustReal no method exceeds all three perception-free controls at any budget.  The order follows what the sequential test has to accumulate: an execution offers about 63 segments on HoloAssist, 14 on CaptainCook4D and 6 on IndustReal (Table~\ref{tab:benchmarks}), and the IndustReal thresholds are set on four validation mistakes.

\noindent\textbf{First and later mistakes.}  On HoloAssist at 0.5 per minute \provia{} recalls twice the share of later mistakes of any baseline and 1.8 times that of any position control (Later in Table~\ref{tab:alarm_test}), and .180 of the first mistakes of an execution against at most .048 for any other row, so its lead there is not confined to the mistakes that follow a corrupted state.

\noindent\textbf{Timing of alarms.}  At 0.5 per minute .82 of \provia{}'s detections on CaptainCook4D and .89 on HoloAssist precede the next step, with a median delay after the end of the mistake of 9.3\,s on CaptainCook4D and 0.5\,s on HoloAssist, where the baselines take 1.0 to 1.6\,s and at most .85 of their detections precede the next step.

\section{Conclusion}
\label{sec:conclusion}

\provia{} reads each step of a procedure against the execution as it actually happened.  It keeps the steps each actor performed apart from the progress it accepts, and raises alarms at a stated false-alarm budget faster than the video arrives.  Scored on complete trials and against controls that use no perception, it ranks mistakes best among the evaluated controlled baselines on four benchmarks spanning cooking, industrial assembly and disassembly, and everyday tasks, over a 23-fold range of step granularity, and removing the record of what was performed at inference costs accuracy on each single-actor benchmark.  As an alarm it recalls the most mistakes at 0.1 false alarms per minute on CaptainCook4D and at every budget on HoloAssist, where its alarms follow the end of the mistake within half a second, earlier than those of any baseline.

\label{sec:end-of-body}
\section*{Acknowledgment}
The project is funded by the Deutsche Forschungsgemeinschaft (DFG, German Research Foundation) -- SFB-1574 -- 471687386.  This work was supported in part by the SmartAge project sponsored by the Carl Zeiss Stiftung (P2019-01-003; 2021-2026).  The authors gratefully acknowledge the computing time provided on the high-performance computer HoreKa by the National High-Performance Computing Center at KIT (NHR@KIT).  This center is jointly supported by the Federal Ministry of Education and Research and the Ministry of Science, Research and the Arts of Baden-W\"urttemberg, as part of the National High-Performance Computing (NHR) joint funding program (\url{https://www.nhr-verein.de/en/our-partners}).  HoreKa is partly funded by the German Research Foundation (DFG).

\bibliographystyle{IEEEtran}
\bibliography{bstctl,reference}

@IEEEtranBSTCTL{IEEEexample:BSTcontrol,
  CTLuse_forced_etal       = "yes",
  CTLmax_names_forced_etal = "6",
  CTLnames_show_etal       = "1"
}

@inproceedings{peddi2024captaincook4d,
  title     = {{CaptainCook4D}: {A} Dataset for Understanding Errors in Procedural Activities},
  author    = {Peddi, Rohith and Arya, Shivvrat and Challa, Bharath and Pallapothula, Likhitha and Vyas, Akshay and Gouripeddi, Bhavya and Zhang, Qifan and Wang, Jikai and Komaragiri, Vasundhara and Ragan, Eric and Ruozzi, Nicholas and Xiang, Yu and Gogate, Vibhav},
  booktitle = {Proc. NeurIPS},
  year      = {2024},
  volume    = {37},
  pages     = {135626--135679}
}

@inproceedings{schoonbeek2024industreal,
  title     = {{IndustReal}: {A} Dataset for Procedure Step Recognition Handling Execution Errors in Egocentric Videos in an Industrial-Like Setting},
  author    = {Schoonbeek, Tim J. and Houben, Tim and Onvlee, Hans and de With, Peter H. N. and van der Sommen, Fons},
  booktitle = {Proc. WACV},
  pages     = {4353--4362},
  year      = {2024}
}

@inproceedings{wang2023holoassist,
  title     = {{HoloAssist}: an Egocentric Human Interaction Dataset for Interactive {AI} Assistants in the Real World},
  author    = {Wang, Xin and Kwon, Taein and Rad, Mahdi and Pan, Bowen and Chakraborty, Ishani and Andrist, Sean and Bohus, Dan and Feniello, Ashley and Tekin, Bugra and Frujeri, Felipe Vieira and Joshi, Neel and Pollefeys, Marc},
  booktitle = {Proc. ICCV},
  pages        = {20270--20281},
  year      = {2023}
}

@inproceedings{flaborea2024prego,
  title     = {{PREGO}: {Online} Mistake Detection in Procedural Egocentric Videos},
  author    = {Flaborea, Alessandro and {D'Amely di Melendugno}, Guido Maria and Plini, Leonardo and Scofano, Luca and {De Matteis}, Edoardo and Furnari, Antonino and Farinella, Giovanni Maria and Galasso, Fabio},
  booktitle = {Proc. CVPR},
  pages     = {18483--18492},
  year      = {2024}
}

@inproceedings{patsch2025mistsense,
  title     = {{MistSense}: {Versatile} Online Detection of Procedural and Execution Mistakes},
  author    = {Patsch, Constantin and Wu, Yuankai and Zakour, Marsil and Salihu, Driton and Steinbach, Eckehard},
  booktitle = {Proc. ICCV},
  pages     = {14528--14537},
  year      = {2025}
}

@article{oquab2023dinov2,
  title={{DINOv2:} {Learning} Robust Visual Features without Supervision},
  author={Oquab, Maxime and Darcet, Timoth{\'e}e and Moutakanni, Th{\'e}o and Vo, Huy V. and Szafraniec, Marc and Khalidov, Vasil and Fernandez, Pierre and Haziza, Daniel and Massa, Francisco and El-Nouby, Alaaeldin and Assran, Mido and Ballas, Nicolas and Galuba, Wojciech and Howes, Russell and Huang, Po-Yao and Li, Shang-Wen and Misra, Ishan and Rabbat, Michael and Sharma, Vasu and Synnaeve, Gabriel and Xu, Hu and Jegou, Herve and Mairal, Julien and Labatut, Patrick and Joulin, Armand and Bojanowski, Piotr},
  journal={Transactions on Machine Learning Research},
  year={2024}
}

@inproceedings{lee2024egoped,
  author    = {Lee, Shih-Po and Lu, Zijia and Zhang, Zekun and Hoai, Minh and Elhamifar, Ehsan},
  title     = {Error Detection in Egocentric Procedural Task Videos},
  booktitle = {Proc. CVPR},
  pages     = {18655--18666},
  year      = {2024},
}

@inproceedings{wang2025amnar,
  author    = {Huang, Wei-Jin and Li, Yuan-Ming and Xia, Zhi-Wei and Tang, Yu-Ming and Lin, Kun-Yu and Hu, Jian-Fang and Zheng, Wei-Shi},
  title     = {Modeling Multiple Normal Action Representations for Error Detection in Procedural Tasks},
  booktitle = {Proc. CVPR},
  pages={27794--27804},
  year      = {2025},
}

@inproceedings{mur2026aem,
  author    = {Guo, Wenliang and Pu, Yujiang and Kong, Yu},
  title     = {Procedural Mistake Detection via Action Effect Modeling},
  booktitle = {Proc. ICLR},
  year      = {2026},
}

@inproceedings{lee2025gtg2vid,
  author    = {Lee, Shih-Po and Elhamifar, Ehsan},
  title     = {Error Recognition in Procedural Videos using Generalized Task Graph},
  booktitle = {Proc. ICCV},
  pages={10009--10021},
  year      = {2025},
}

@inproceedings{lee2026axg,
  author    = {Lee, Shih-Po and Elhamifar, Ehsan},
  title     = {{AXG-Reasoner}: {Error} Detection and Explanation in Long Task Videos with Vision--Language Models},
  booktitle = {Proc. CVPR},
  pages = {3421--3431},
  year      = {2026},
}

@article{seminara2606zeprom,
  author  = {Ozsoy, Serdar and Doorenbos, Lars and Spurio, Federico and Francesca, Gianpiero and Gall, Juergen},
  title   = {The Unreasonable Effectiveness of {VLMs} for Zero-shot Procedural Mistake Detection},
  journal = {arXiv preprint arXiv:2606.21579},
  year    = {2026},
}

@article{mistexit2026,
  author  = {Majumder, Sagnik and Nethi, Anish and Al-Halah, Ziad and Grauman, Kristen},
  title   = {{MistExit}: Learning to Exit for Early Mistake Detection in Procedural Videos},
  journal = {arXiv preprint arXiv:2603.14252},
  year    = {2026},
}

@article{flaborea2024tiprego,
  title={{TI-PREGO:} {Chain} of Thought and In-Context Learning for Online Mistake Detection in PRocedural EGOcentric Videos},
  author={Plini, Leonardo and Scofano, Luca and {De Matteis}, Edoardo and {D'Amely di Melendugno}, Guido Maria and Flaborea, Alessandro and Sanchietti, Andrea and Farinella, Giovanni Maria and Galasso, Fabio and Furnari, Antonino},
  journal={Computer Vision and Image Understanding},
  volume={264},
  pages={104613},
  year={2026},
  publisher={Elsevier}
}

@inproceedings{seminara2024dtgl,
  author    = {Seminara, Luigi and Farinella, Giovanni Maria and Furnari, Antonino},
  title     = {Differentiable Task Graph Learning: Procedural Activity Representation and Online Mistake Detection from Egocentric Videos},
  booktitle = {Proc. NeurIPS},
  volume    = {37},
  pages     = {59373--59407},
  year      = {2024},
}

@article{schoonbeek2025stormpsr,
  author  = {Schoonbeek, Tim J. and Hung, Shao-Hsuan and Lehman, Dan and Onvlee, Hans and Kustra, Jacek and {de With}, Peter H. N. and {van der Sommen}, Fons},
  title   = {Learning to Recognize Correctly Completed Procedure Steps in Egocentric Assembly Videos through Spatio-Temporal Modeling},
  journal = {Computer Vision and Image Understanding},
  volume  = {262},
  pages   = {104528},
  year    = {2025},
}

@inproceedings{wen2026impact,
  title={{IMPACT:} {A} Dataset for Multi-Granularity Human Procedural Action Understanding in Industrial Assembly},
  author={Wen, Di and Zhong, Zeyun and Schneider, David and Zaremski, Manuel and Kunzmann, Linus and Shi, Yitian and Liu, Ruiping and Chen, Yufan and Zheng, Junwei and Li, Jiahang and Hemmerich, Jonas and Tong, Qiyi and Grauberger, Patric and Ajoudani, Arash and Paudel, Danda Pani and Matthiesen, Sven and Deml, Barbara and Beyerer, J{\"u}rgen and {Van Gool}, Luc and Stiefelhagen, Rainer and Peng, Kunyu},
  booktitle={Proc. MM},
  year={2026}
}

@inproceedings{liu2026guideme,
  title={{GuideMe:} {Multi-domain} Task Guidance and Intervention in Streaming Video},
  author={Liu, Fang and Chen, Jinpeng and Xu, Ke and Liu, Yuhao and Guan, Huankang and Lu, Xudong and Yang, Bo and Hancke, Gerhard and Liu, Rui and Lau, Rynson W. H.},
  booktitle={Proc. ECCV},
  year={2026}
}

@article{kundu2026pwr,
  title={Plan, Watch, Recover: A Benchmark and Architectures for Proactive Procedural Assistance},
  author={Kundu, Kaustav and Shrivastava, Ritvik and Arap, Maxim and Wang, Nanshu and Zhu, Xianhui and Fettes, Quintin and Tiwari, Gautam and Suresh, Parth and Moutakanni, Th{\'e}o and {Castillejo Munoz}, Alejandro and Bolourchi, Allen and Fung, Pascale and Donmez, Pinar and Damavandi, Babak and Kumar, Anuj and Moon, Seungwhan},
  journal={arXiv preprint arXiv:2606.04970},
  year={2026}
}

@article{shiryaev1963optimum,
  title={On optimum methods in quickest detection problems},
  author={Shiryaev, Albert N.},
  journal={Theory of Probability \& Its Applications},
  volume={8}, number={1}, pages={22--46}, year={1963}
}

@article{roberts1966comparison,
  title={A comparison of some control chart procedures},
  author={Roberts, S. W.},
  journal={Technometrics},
  volume={8}, number={3}, pages={411--430}, year={1966}
}

@article{ding2023every,
  title={Every Mistake Counts in Assembly},
  author={Ding, Guodong and Sener, Fadime and Ma, Shugao and Yao, Angela},
  journal={arXiv preprint arXiv:2307.16453},
  year={2023}
}

@inproceedings{sener2022assembly101,
  title     = {{Assembly101}: {A} Large-Scale Multi-View Video Dataset for Understanding Procedural Activities},
  author    = {Sener, Fadime and Chatterjee, Dibyadip and Shelepov, Daniel and He, Kun and Singhania, Dipika and Wang, Robert and Yao, Angela},
  booktitle = {Proc. CVPR},
  pages={21096--21106},
  year      = {2022}
}

@inproceedings{an2023miniroad,
  title     = {{MiniROAD:} {Minimal} {RNN} Framework for Online Action Detection},
  author    = {An, Joungbin and Kang, Hyolim and Han, Su Ho and Yang, Ming-Hsuan and Kim, Seon Joo},
  booktitle = {Proc. ICCV},
  pages={10341--10350},
  year      = {2023}
}

@inproceedings{stolcke1992hmm,
  title={Hidden {M}arkov Model Induction by {B}ayesian Model Merging},
  author={Stolcke, Andreas and Omohundro, Stephen},
  booktitle={Proc. NeurIPS},
  volume={5},
  pages={11--18},
  year={1992}
}

@article{zhang2026impacthoi,
  title={{IMPACT-HOI:} {Supervisory} Control for Onset-Anchored Partial {HOI} Event Construction},
  author={Zhang, Haoshen and Wen, Di and Peng, Kunyu and Schneider, David and Zhong, Zeyun and Jaus, Alexander and Marinov, Zdravko and Wei, Jiale and Liu, Ruiping and Zheng, Junwei and Chen, Yufan and Zhang, Yufeng and Luo, Yuanhao and Qi, Lei and Stiefelhagen, Rainer},
  journal={arXiv preprint arXiv:2605.01666},
  year={2026}
}

@inproceedings{peng2024openset,
  title     = {Navigating Open Set Scenarios for Skeleton-Based Action Recognition},
  author    = {Peng, Kunyu and Yin, Cheng and Zheng, Junwei and Liu, Ruiping and Schneider, David and Zhang, Jiaming and Yang, Kailun and Sarfraz, M. Saquib and Stiefelhagen, Rainer and Roitberg, Alina},
  booktitle = {Proc. AAAI},
  pages={4487--4496},
  year      = {2024}
}
\end{document}